\documentclass[journal]{IEEEtran}
\usepackage{makecell}
\usepackage{booktabs}
\usepackage[utf8]{inputenc}
\usepackage{siunitx}
\usepackage[colorlinks=true, allcolors=blue]{hyperref}
\usepackage{amsmath}
\usepackage{amsfonts}
\usepackage{soul} 
\usepackage{subcaption}
\usepackage{stfloats}
\usepackage{booktabs}
\usepackage{enumitem}
\usepackage{cite} 

\usepackage{tikz,xcolor}

\usepackage[ruled,linesnumbered]{algorithm2e}
\usepackage{algpseudocode}

\definecolor{lime}{HTML}{A6CE39}
\DeclareRobustCommand{\orcidicon}
{
\begin{tikzpicture}
\draw[lime, fill=lime] (0,0) circle [radius=0.16]
 node[white] {{\fontfamily{qag}\selectfont \tiny ID}};\draw[white, fill=white] (-0.0625,0.095) circle [radius=0.007];
 \end{tikzpicture}
 \hspace{0mm}}
\foreach \x in {A, ..., Z}{%
 \expandafter\xdef\csname orcid\x\endcsname{\noexpand\href{https://orcid.org/\csname orcidauthor\x\endcsname}{\noexpand\orcidicon}}
}

\setlist[enumerate]{itemsep = 0pt, parsep = 0pt, topsep = 0pt} 
\setlist[itemize]{itemsep = 0pt, parsep = 0pt, topsep = 0pt} 

\begin{document}

\title{Data Shift of Object Detection in Autonomous Driving}

\author{Lida Xu \vspace*{-30pt}

\thanks{Lida Xu is with the Department of Computer Science and Engineering at Southern University of Science and Technology, Shenzhen, China.}
}

\markboth{Journal of \LaTeX\ Class Files,~Vol.
}%
{Shell \MakeLowercase{\textit{et al.}}: Bare Demo of IEEEtran.cls for IEEE Journals}

\maketitle

\begin{abstract}
With the widespread adoption of machine learning technologies in autonomous driving systems, their role in addressing complex environmental perception challenges has become increasingly crucial. However, existing machine learning models exhibit significant vulnerability, as their performance critically depends on the fundamental assumption that training and testing data satisfy the independent and identically distributed condition, which is difficult to guarantee in real-world applications. Dynamic variations in data distribution caused by seasonal changes, weather fluctuations lead to data shift problems in autonomous driving systems.
This study investigates the data shift problem in autonomous driving object detection tasks, systematically analyzing its complexity and diverse manifestations. We conduct a comprehensive review of data shift detection methods and employ shift detection analysis techniques to perform dataset categorization and balancing. Building upon this foundation, we construct an object detection model. 
To validate our approach, we optimize the model by integrating CycleGAN-based data augmentation techniques with the YOLOv5 framework. Experimental results demonstrate that our method achieves superior performance compared to baseline models on the BDD100K dataset.

\end{abstract}


\IEEEpeerreviewmaketitle

\section{Introduction}

In recent years, autonomous driving technology has undergone rapid development and its application has become increasingly widespread. 
This technology integrates artificial intelligence with sensors, localization, and control techniques to establish a comprehensive system for perception, decision-making, and control. 
It utilizes sensors to acquire information about the surrounding environment, vehicle status, and location. After information fusion, a computational unit makes behavioral decisions and plans routes, and the control system manipulates the vehicle based on these decisions and commands to achieve assisted driving functionalities. 
The application of this technology not only reduces the driver's operational burden and enhances driving comfort but also holds the promise of playing a key role in improving traffic efficiency and reducing the incidence of traffic accidents.

Within the autonomous driving framework, the environmental perception system is a core component, and the performance of object detection, as a critical part of this system, directly impacts the safety and robustness of the entire system. 
With the continuous advancement of deep learning \cite{lan2022vision}, the accuracy of object detection models has reached a high level in ideal test environments. 
However, in complex and variable real-world dynamic scenes, a distributional gap often exists between the training data and the actual environment, a problem known as data shift. This shift can lead to insufficient generalization capability of the model, causing a significant decline in object detection accuracy and thereby creating safety hazards.

To address the safety challenges posed by data shift and mitigate the performance degradation caused by this distributional discrepancy, data augmentation has emerged as a crucial technique for enhancing model generalization. By systematically transforming original data (e.g., through image rotation), data augmentation expands data diversity, simulates variations in real-world scenes, helps the model learn robust features, alleviates performance decay from distribution gaps, and improves generalization.

This paper uses data shift detection as a starting point to propose an optimized data augmentation scheme based on CycleGAN~\cite{zhu2017unpaired}, and its effectiveness is experimentally validated on the BDD100K dataset~\cite{yu2020bdd100k}.
Through the optimization of the object detection model, a significant improvement in detection performance was ultimately achieved. The specific research contents are as follows:
\begin{enumerate}
\item Data Shift Detection and Analysis: Applying data shift detection methods to the dataset to extract and analyze its distribution characteristics and other information, preparing for subsequent optimization work.
\item Optimization Method Based on YOLOv5 and Data Augmentation: Using YOLOv5~\cite{yolo} as the base detection framework, an improved model is proposed by incorporating data augmentation strategies.
\item Experimental Validation of Algorithm Feasibility: Establishing an experimental environment using the BDD100K dataset to test the proposed method.
\end{enumerate}

The experimental results demonstrate that the proposed optimization method enhances the performance of object detection, validating the algorithm's feasibility and practicality. 
The optimization scheme presented in this paper is of great significance for enhancing the robustness and safety of autonomous driving systems under varying environmental conditions and data distribution changes.

\section{Related Work}
\label{sec:related_work}

The reliability of object detection technology is directly linked to the safety of autonomous driving, and data shift presents an unavoidable challenge in this domain. 
Therefore, mitigating the effects of data shift and selecting appropriate object detection techniques are crucial for enhancing autonomous driving safety.

As a core area of computer vision, research on object detection has primarily focused on improving accuracy, speed, robustness, and adaptability to various scenes \cite{lan2019evolving}. 
Traditional machine learning-based object detection methods \cite{lan2022class} rely on hand-crafted feature extractors (e.g., SIFT~\cite{lowe2004distinctive}, HOG~\cite{dalal2005histograms}) combined with classifiers like Support Vector Machines for object localization and recognition. 
While effective in simple scenarios, these methods suffer from limited generalization capabilities and accuracy, struggling to handle variations in data distribution.

Two-stage object detection algorithms based on Convolutional Neural Networks (CNNs), such as R-CNN~\cite{girshick2014rich}, Fast R-CNN~\cite{girshick2015fast}, Faster R-CNN~\cite{ren2015faster}, SPP-Net~\cite{he2015spatial}, and FPN~\cite{lin2017feature}, first generate region proposals and then perform classification and regression. Although they achieve high accuracy, the need to process a large number of proposals makes them too slow to meet the real-time requirements of autonomous driving \cite{lan2024dir}. 
In contrast, one-stage algorithms like the SSD~\cite{liu2016ssd} and YOLO series~\cite{redmon2016you} perform predictions in a single forward pass, reducing computational complexity and striking a balance between speed and accuracy, thereby satisfying the real-time demands of autonomous systems. 
More recently, novel detectors based on the Transformer architecture~\cite{vaswani2017attention}, such as DETR~\cite{carion2020end}, leverage self-attention mechanisms to capture global dependencies, showing new potential for handling complex scenes and improving generalization. Despite the distinct advantages of each architecture, there is a consensus that no single model can fully address the generalization problem in the real world, making active data-side intervention an indispensable step.

Methods for detecting data shift have evolved from relying on external statistics to analyzing internal model features. Early approaches applied statistical tests directly to high-dimensional data but were often hampered by the "curse of dimensionality." To overcome this, research shifted towards two-sample testing on lower-dimensional representations of the data. In this direction, the work by Rabanser et al.~\cite{rabanser2019failing} systematically demonstrated the effectiveness of using a model's own outputs (e.g., prediction distributions) as a means of dimensionality reduction, which is the core idea behind methods like BBSD. This approach effectively combines statistical tests with a model's internal information, significantly improving detection efficiency and the ability to handle high-dimensional data.

Meanwhile, other technical paths have also been explored. For instance, monitoring performance metrics like model accuracy can indicate a shift, but this method suffers from a time lag. Another approach involves analyzing deeper internal model information, such as intermediate layer features or reconstruction errors. While more sensitive, this method is often dependent on specific model architectures. Furthermore, out-of-distribution (OOD) detectors~\cite{hendrycks2016baseline} that act as a post-processing "circuit breaker" have a logical flaw, as their reactive nature cannot proactively guarantee safety.

The primary goal of data shift mitigation strategies is to address the distributional mismatch between training data and data encountered in real-world deployment scenarios. To alleviate the performance degradation caused by this discrepancy, data augmentation has become a key technique for improving model generalization. Traditional augmentation methods, such as rotation, scaling, flipping, and adding noise, generate new samples through simple geometric transformations or perturbations.~\cite{shorten2019survey} Although easy to implement and capable of providing some generalization benefits, they are limited by a lack of sample diversity and semantic information, making them insufficient for simulating complex real-world variations.

In recent years, generative augmentation techniques, particularly those based on Diffusion~\cite{ho2020denoising} and GAN~\cite{goodfellow2014generative} models, have advanced significantly. Diffusion models have achieved breakthroughs in the quality of image synthesis, but their high computational cost and challenges in maintaining controllable and precise geometric structures remain obstacles. 
Concurrently, Generative Adversarial Networks (GANs) have offered a new path for data augmentation. 
Through the adversarial interplay between a generator and a discriminator, GANs can learn the underlying distribution of real data and generate highly realistic new samples. 
CycleGAN (Cycle-Consistent Generative Adversarial Network)~\cite{zhu2017unpaired}, in particular, has become a research hotspot due to its ability to perform cross-domain image translation without paired training data. 
However, unconstrained generative models like CycleGAN can introduce semantically incorrect data, which may harm rather than improve model performance.

Based on the analysis above, the limitations of using these methods in isolation can be summarized as follows:
\begin{enumerate}
\item Vulnerability of Object Detectors: Object detection models are fragile and experience significant performance drops when faced with data shifts.
\item Unreliability of Passive Detection: Reactive or passive shift detection methods are unreliable as they identify problems after they have already occurred, which is insufficient for safety-critical systems.
\item Semantic Risks of Generative Augmentation: Unconstrained generative augmentation techniques, such as a standard CycleGAN, risk introducing samples with semantic errors, potentially degrading model robustness.
\end{enumerate}

The CycleGAN-based data augmentation scheme proposed in this paper uses data shift detection as a starting point. 
It leverages CycleGAN to generate data that more closely aligns with real-world scenarios, thereby bridging the distributional gap between training and deployment. 
This approach provides a more comprehensive training basis for the object detection model. 
By combining data shift handling with a targeted exploration of object detection technology, our solution not only mitigates the impact of data shift but also effectively enhances the robustness of the detector.

\section{Methodology}
\label{sec:methodology}

In autonomous driving scenarios, environmental factors such as lighting and weather can cause a distribution shift between training data and real-world application data, significantly impacting the performance of object detection models \cite{xu2019online}. 
This problem is particularly pronounced when daytime data is abundant while nighttime data is relatively sparse, limiting the model's generalization ability in nocturnal scenes.
To address this issue, this paper proposes an object detection optimization system based on CycleGAN for data augmentation \cite{lan2024sustechgan}. 
The system aims to mitigate the shift between datasets through style transfer, thereby enhancing the model's detection performance in the target domain (i.e., nighttime scenes).

\subsection{System Architecture and Experimental Design}
\begin{figure}[!ht] \centering
 \includegraphics[width=0.9\linewidth, trim={5 15 5 15}, clip]{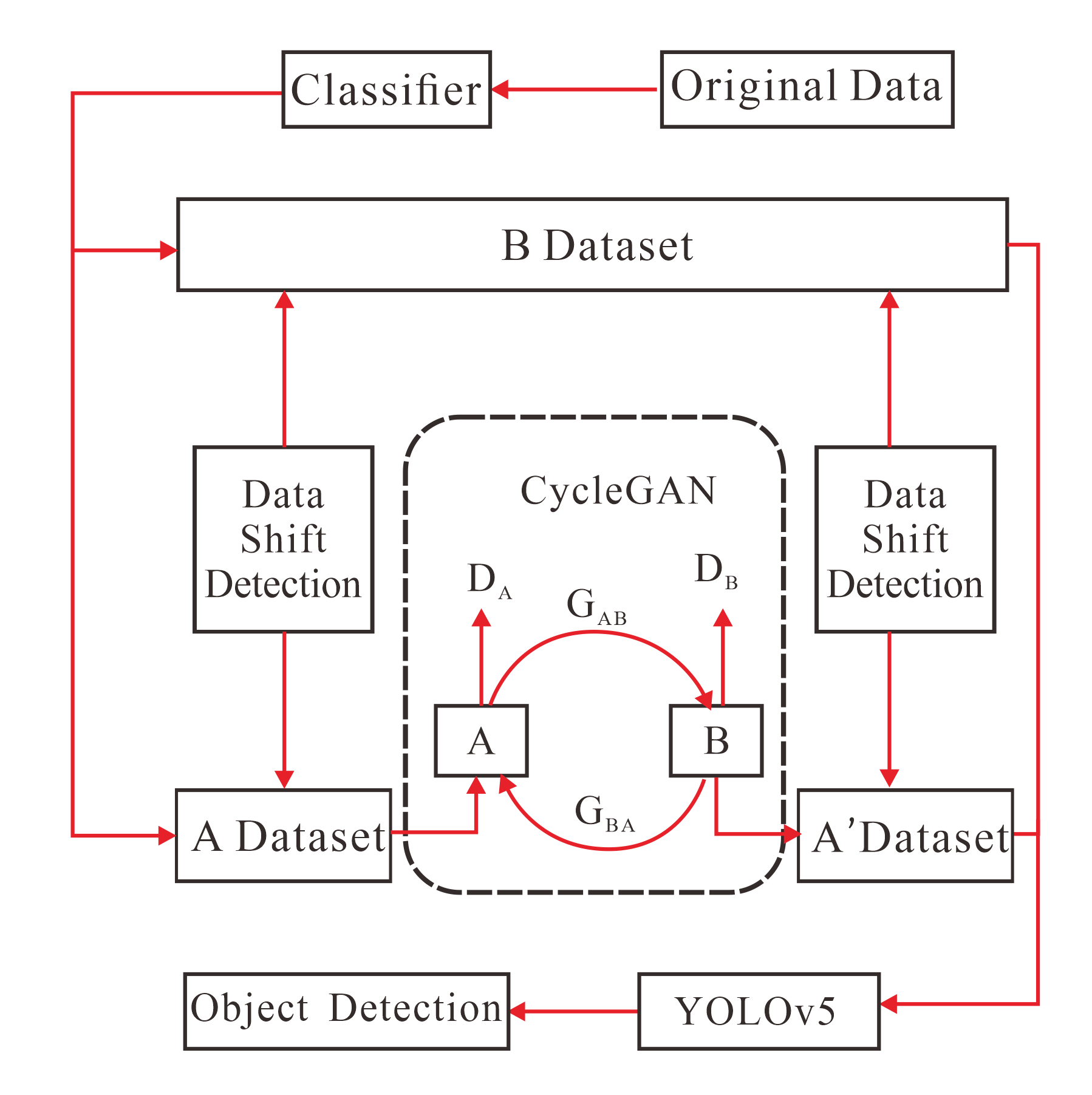}
\caption{The proposed system architecture for object detection optimization based on data shift detection and CycleGAN.}
 \label{fig:system_architecture}
\end{figure}

\subsubsection{Experimental Procedure}
\begin{enumerate}
\item Dataset Preparation and Partitioning: The BDD100K dataset is used as the base dataset. Using image metadata or a classifier, the dataset is partitioned into a source domain (daytime images), denoted as $D_A$, and a target domain (nighttime images), denoted as $D_B$.

\item Baseline Model Training and Evaluation ($M_0$): A standard YOLOv5s model is trained as the baseline detector. This model, $M_0$, is trained on several datasets with varying ratios of day-to-night images, including 0N\_40000D (0 night, 40,000 day images), 5000N\_35000D, 10000N\_30000D, 15000N\_25000D, and 20000N\_20000D. The performance of the baseline model is then evaluated on a validation set composed exclusively of nighttime images.

\item Initial Data Shift Detection: The data shift detection method described in the previous section is applied to quantify the distributional discrepancy between the source domain $D_A$ and target domain $D_B$. This serves as a quantitative baseline for the initial shift.

\item CycleGAN Model Training: To address the identified data shift, a CycleGAN model \cite{zhu2017unpaired} is trained using the partitioned daytime dataset $D_A$ and nighttime dataset $D_B$.

\item CycleGAN-based Data Augmentation: The trained CycleGAN is used to perform style transfer, converting daytime images from the training set into nighttime-style images. These newly generated images are then combined with the original real nighttime images to form an augmented training set, $D_{\text{train}}^{\text{aug}}$.

\item Second Data Shift Detection: After data augmentation, shift detection is performed again to compare the distribution of the augmented training set $D_{\text{train}}^{\text{aug}}$ with the target nighttime domain. A reduction in the distribution difference is expected.

 \item Optimized Model Training and Evaluation ($M_1$): The YOLOv5s model is retrained or fine-tuned on the augmented dataset $D_{\text{train}}^{\text{aug}}$ to produce the optimized model, $M_1$. Its performance is evaluated on the same pure-nighttime validation set used for the baseline model.

\item Comparative Analysis: The final phase involves a comprehensive comparison between the optimized model $M_1$ and the baseline model $M_0$. The primary metric for comparison is the mean Average Precision (mAP) on the nighttime test set. This analysis aims to validate the effectiveness of the proposed strategy and to explore the potential correlation between the quantified data shift and the improvement in detection performance.
\end{enumerate}

\subsection{Principles and Mathematical Formulation}

\subsubsection{Dataset Definitions}
The datasets used in this study are formally defined as follows:
Let $D_{\text{BDD100K}}$ be the complete BDD100K dataset. From this, we partition a source domain dataset (daytime) $D_A = \{(x_i^A, y_i^A)\}_{i=1}^{N_A}$ and a target domain dataset (nighttime) $D_B = \{(x_j^B, y_j^B)\}_{j=1}^{N_B}$, where $N_A$ and $N_B$ are the number of day and night images, respectively. The initial training set for the baseline model is $D_{\text{train}}^{\text{base}}$, constructed with various day/night ratios. A pure nighttime validation set, $D_{\text{val\_night}}$, is also defined. For training CycleGAN, subsets $D_{\text{train\_A\_for\_cyclegan}} \subset D_A$ and $D_{\text{train\_B\_for\_cyclegan}} \subset D_B$ are used.

\subsubsection{Baseline YOLOv5s Model Training ($M_0$)}
\begin{figure}[!ht]
\centering
\includegraphics[width=0.98\linewidth,trim={0 15 0 20},clip]{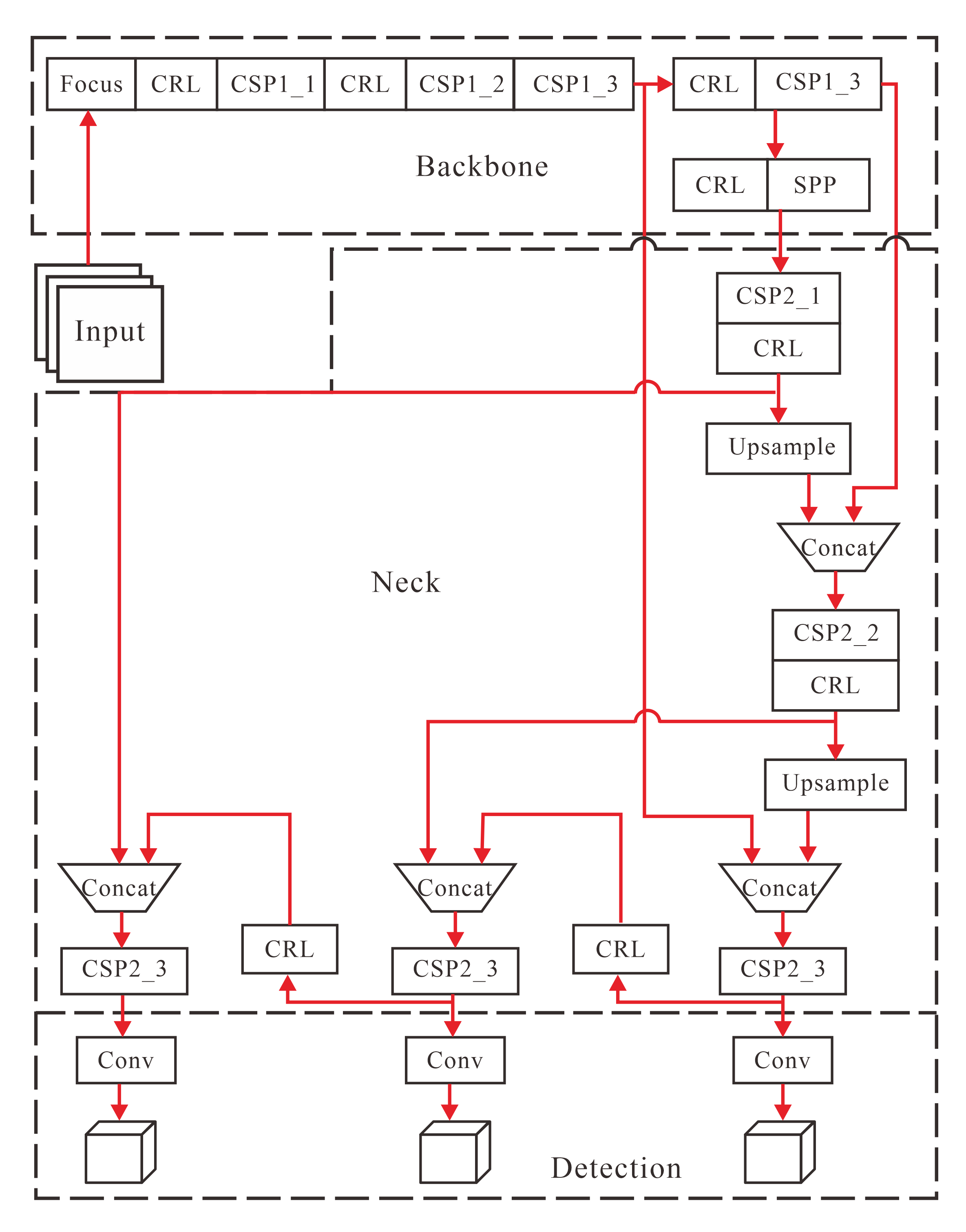}
\caption{YOLOv5s Architecture Diagram}
\label{yolov5s_architecture}
\end{figure}
The core architecture of YOLOv5 is as follows: Input, Backbone, Neck, and Head. Figure \ref{yolov5s_architecture} shows the architecture diagram of YOLOv5s.

The baseline model $M_0$, with parameters $\theta_0$, is trained by minimizing the standard YOLOv5 loss, $L_{\text{YOLO}}(x, y, \theta)$, over the initial training set $D_{\text{train}}^{\text{base}}$:
\begin{equation}
\theta_0 = \arg\min_{\theta} \sum_{(x_k, y_k) \in D_{\text{train}}^{\text{base}}} L_{\text{YOLO}}(x_k, y_k, \theta)
\end{equation}
After training, its performance, $\text{mAP}(M_0, D_{\text{val\_night}})$, is calculated on the nighttime validation set.

\subsubsection{CycleGAN Model for Data Augmentation}
In the GAN framework, the generator G and the discriminator D are trained adversarially by constructing a minimax game. The core objective is to approximate the true probability distribution of the training data through this dynamic game. The value function $V(D, G)$ is defined as:
\begin{align}
\label{eq:gan_objective}
\min_{G} \max_{D} V(D, G) = & \mathbb{E}_{x \sim p_{\text{data}}(x)}[\log D(x)] \nonumber \\
& + \mathbb{E}_{z \sim p_{z}(z)}[\log(1 - D(G(z)))]
\end{align}

In this game mechanism, the discriminator D aims to maximize the probability of correctly classifying real samples and generated samples—that is, maximizing $V(D, G)$—thereby enhancing its ability to distinguish between real and fake data. Conversely, the generator G aims to minimize $V(D, G)$, forcing itself to produce counterfeit samples that are convincing enough to deceive the discriminator. At the theoretical equilibrium, the generator G will be able to produce samples whose statistical properties are identical to the real data distribution. At this point, the discriminator D's probability of correctly identifying both real and generated samples converges to 0.5, indicating that the generative model has successfully approximated the underlying data distribution.

\begin{figure}[!ht]
\centering
\includegraphics[width=0.98\linewidth,trim={5 5 5 5},clip]{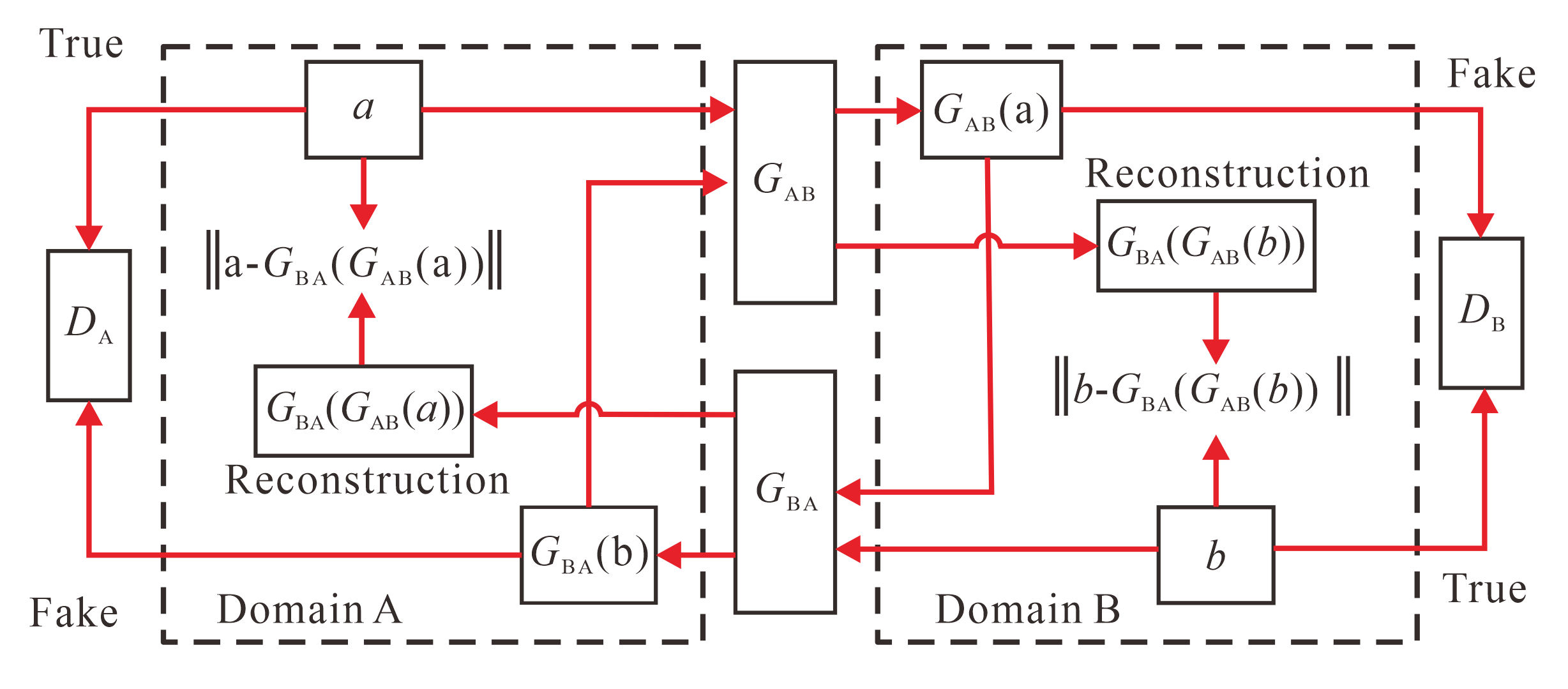}
 \caption{The architecture of CycleGAN, showing the two generators and two discriminators for unpaired image-to-image translation.}
\label{fig:cyclegan_architecture}
\end{figure}

The total loss $L(G_{AB}, G_{BA}, D_A, D_B)$ is a weighted sum of these losses. The overall training objective is a minimax problem:
\begin{equation}
G_{AB}^*, G_{BA}^* = \arg\min_{G_{AB}, G_{BA}} \max_{D_A, D_B} L(G_{AB}, G_{BA}, D_A, D_B)
\end{equation}
Once the generator $G_{AB}^*$ is trained, it is used to create a set of generated nighttime images, $D_{\text{generated\_night}}$. The final augmented training set is constructed as:
\begin{equation}
D_{\text{train}}^{\text{aug}} = D_{\text{real\_night}} \cup D_{\text{generated\_night}}
\end{equation}
where $D_{\text{real\_night}}$ is the set of real nighttime images in the training split.

\subsubsection{Data Shift Quantification}
In machine learning, the discrepancy between the distribution of training data and that of the actual test data is known as data shift \cite{moreno2012unifying}. The three types of shift are summarized in Table \ref{tab:dataset-shift}.\cite{moreno2012unifying}:
\begin{table}[!ht]
\centering
\caption{Three Types of Dataset Shift}
\label{tab:dataset-shift}
\begin{tabular}{l|l}
\toprule
\textbf{Shift Type} & \textbf{Definition} \\
\midrule
Covariate Shift & $P_{src}(x) \ne P_{tgt}(x)$ but $P_{src}(y|x) = P_{tgt}(y|x)$ \\
\midrule
Label Shift  & $P_{src}(y) \ne P_{tgt}(y)$ but $P_{src}(x|y) = P_{tgt}(x|y)$ \\
\midrule
Concept Shift & $P_{src}(y|x) \ne P_{tgt}(y|x)$ but $P_{src}(x) = P_{tgt}(x)$ \\
\bottomrule
\end{tabular}
\end{table}

The data shift detection method used in this experiment mainly involves two key stages: dimensionality reduction and a two-sample distribution test.
Figure \ref{rd_pipeline} shows the flowchart for detecting dataset differences. The specific process is as follows:
\begin{figure}[!ht]
\centering
 \includegraphics[width=0.8\linewidth]{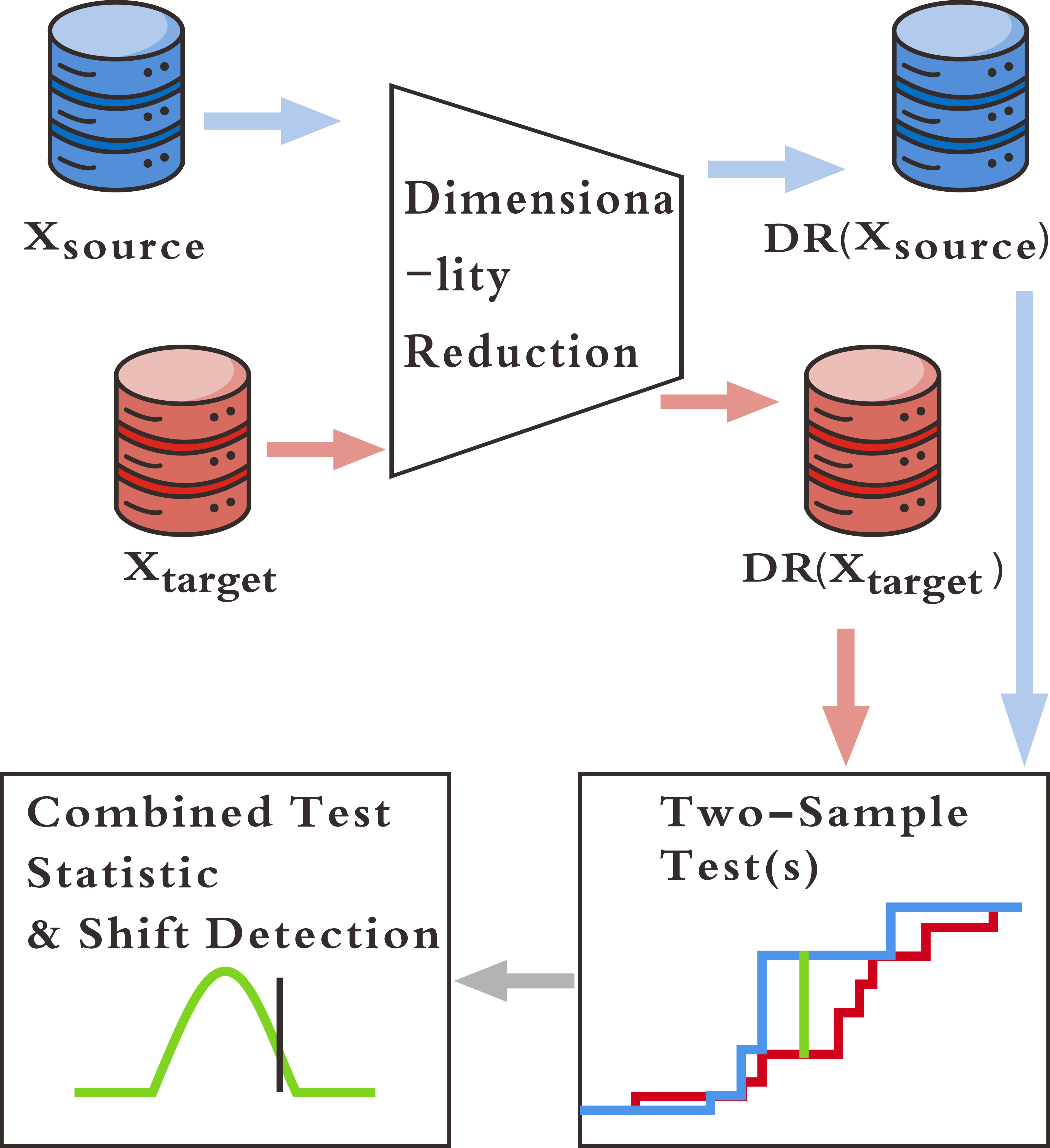}
 \caption{A pipeline for dataset shift detection using dimensionality reduction.}
 \label{rd_pipeline}
\end{figure}
PCA~\cite{PCA} is used as a feature extractor, $f_{\text{embed}}$, to obtain low-dimensional representations of the image sets:

\begin{itemize}
    \item Features of the baseline training set:
    \[ Z_{\text{base}} = \{f_{\text{embed}}(x_k) \mid x_k \in D_{\text{train}}^{\text{base}}\} \]
    \item Features of the augmented training set:
    \[ Z_{\text{aug}} = \{f_{\text{embed}}(x_k) \mid x_k \in D_{\text{train}}^{\text{aug}}\} \]
    \item Features of the nighttime validation set:
    \[ Z_{\text{val\_night}} = \{f_{\text{embed}}(x_k) \mid x_k \in D_{\text{val\_night}}\} \]
\end{itemize}

Afterwards, two-sample test methods are applied to the dimensionally-reduced data to compare the distributions of the source and target data. We then use MMD test \cite{MMD} to compute the shift score between the baseline training set and the nighttime validation set, $S_{\text{drift\_base}} = \text{MMD}(Z_{\text{base}}, Z_{\text{val\_night}})$, and between the augmented training set and the nighttime validation set, $S_{\text{drift\_aug}} = \text{MMD}(Z_{\text{aug}}, Z_{\text{val\_night}})$. Finally, a combined test statistic is used to determine whether a significant difference exists between the two datasets. We hypothesize that $S_{\text{drift\_aug}} < S_{\text{drift\_base}}$.

\subsubsection{Optimized YOLOv5s Model Training ($M_1$)}
The optimized model $M_1$, with parameters $\theta_1$, is trained on the augmented dataset $D_{\text{train}}^{\text{aug}}$. The training objective is to minimize the total loss on this new dataset:
\begin{equation}
\theta_1 = \arg\min_{\theta} \sum_{(x_k, y_k) \in D_{\text{train}}^{\text{aug}}} L_{\text{YOLO}}(x_k, y_k, \theta)
\end{equation}

\subsubsection{Performance Evaluation and Comparison}
The final stage is a comprehensive evaluation. The performance of the optimized model $M_1$ will be evaluated on both a pure nighttime test set, $D_{\text{test\_night}}$, and a pure daytime test set, $D_{\text{test\_day}}$. The core comparison will be between $\text{mAP}(M_1, D_{\text{test\_night}})$ and $\text{mAP}(M_0, D_{\text{test\_night}})$ to quantify the improvement in the target domain. We will also compare performance on the source domain, $\text{mAP}(M_1, D_{\text{test\_day}})$ versus $\text{mAP}(M_0, D_{\text{test\_day}})$, to assess any potential negative impact. Additionally, the analysis will investigate the relationship between the quantified shift scores ($S_{\text{drift\_base}}$ and $S_{\text{drift\_aug}}$) and the final detection performance of models $M_0$ and $M_1$.

\section{Experiments}
\label{sec:experiments}

This section details the series of experiments conducted to investigate the problem of data shift in object detection for autonomous driving. First, we describe the basic experimental setup, including the hardware and software configurations, dataset details, and baseline model parameters. Subsequently, we elaborate on the experiments designed to analyze the impact of data shift, which include preliminary tests to validate the effectiveness of our data shift detection method, experiments for detecting and quantifying data shift between day and night scenes, and an evaluation of the effect of data shift on the performance of the YOLOv5s model. 
Finally, we introduce the model optimization strategy adopted to mitigate the impact of data shift and the design of our comparative experiments.

\subsection{Experimental Setup}

\subsubsection{Experimental Environment}
The hardware and software environment used for our experiments is detailed in \autoref{tab:experimental_environment}.

\begin{table}[!ht]
    \centering
    \caption{Experimental Environment.}
    \label{tab:experimental_environment}
    \begin{tabular}{l|l}
        \toprule
        \multicolumn{2}{c}{\textbf{Experimental Environment}} \\
        \midrule
        GPU & NVIDIA GeForce RTX 4090 (24GB) \\
        \midrule
        CPU & 16 vCPU Intel(R) Xeon(R) Gold 6430 \\
        \midrule
        Memory & 120 GB \\
        \midrule
        Operating System & Ubuntu 20.04 \\
        \midrule
        CUDA Version & 11.3 \\
        \midrule
        Python Version & 3.8 \\
        \midrule
        PyTorch Version & 1.10.0 \\
        \midrule
        YOLOv5 Version & 6.0 \\
        \bottomrule
    \end{tabular}
\end{table}

In this study, the training parameters for the YOLOv5s object detection model were set as follows: `batch-size` was set to 128, the initial learning rate was 0.01, and the Adam optimizer was used for loss optimization. 
The model was trained for 20 epochs. All experiments were conducted with the same random seed to ensure reproducibility.

\subsubsection{Dataset and Model Configuration}
The core dataset used in this study is BDD100K \cite{yu2020bdd100k}. BDD100K is a large-scale, diverse dataset for autonomous driving, containing 100,000 images with a resolution of $1280 \times 720$, covering a wide range of weather conditions, times of day, and scenes. While the dataset is annotated for 10 different tasks, our work focuses on the road object detection task, which includes 10 object classes (e.g., car, truck, person, traffic light).

To study the effects of data shift, we primarily focused on the daytime and nighttime images within the BDD100K dataset. The data was partitioned as follows:

\textbf{Training Set Construction.} To simulate varying degrees of data shift, we constructed multiple training subsets by combining daytime and nighttime images from the 70,000 training images of BDD100K in different proportions. As shown in \autoref{tab:training_set_proportion}, we created five training sets with different day-to-night image ratios (D for Day, N for Night), maintaining a total of 40,000 images each.

\begin{table}[!ht]
    \centering
    \small
    \caption{Construction of Training Sets with Different Day/Night Ratios.}
    \label{tab:training_set_proportion}
    \begin{tabular}{l|l}
        \toprule
        \textbf{Training Set \& Shift Severity} & \textbf{Day/Night Image Ratio (D/N)} \\
        \midrule
        Set A (Severe Daytime Shift) & 40000D / 0N \\
        \midrule
        Set B (Moderate Daytime Shift) & 35000D / 5000N \\
        \midrule
        Set C (Mild Daytime Shift) & 30000D / 10000N \\
        \midrule
        Set D (Slight Daytime Shift) & 25000D / 15000N \\
        \midrule
        Set E (Balanced Day/Night) & 20000D / 20000N \\
        \bottomrule
    \end{tabular}
\end{table}

\textbf{Validation and Test Sets.} From the original BDD100K validation set (10,000 images), we selected only nighttime images to construct a night-only validation set (denoted as $Val_{night}$) and a night-only test set (denoted as $Test_{night}$). These sets were used to evaluate the model's performance in the target domain (nighttime).

This study employs YOLOv5s as the baseline object detection model architecture.

\subsubsection{Data Shift Detection Setup}
\label{subsubsec:stat_test_setup}
In our data shift detection experiments, we utilized the Maximum Mean Discrepancy (MMD) as a two-sample statistical test to compare the distribution differences between two datasets (e.g., source domain features vs. target domain features). For each test, we randomly sampled 1,000 instances from each of the two datasets and computed the MMD statistic and its corresponding $p$-value. A $p$-value below the significance level of $\alpha = 0.05$ was interpreted as evidence of a significant distribution shift. To ensure the stability of our results, each MMD test was repeated 30 times, and the average $p$-value was reported.

\subsubsection{CycleGAN Training Setup}
To perform image style transfer from day to night and vice versa, we employed a CycleGAN model. The model was trained on a specific subset of the BDD100K dataset, comprising 4,386 nighttime images as the target domain (Night) and 8,284 daytime images as the source domain (Day). The training was conducted for a total of 150 epochs. The detailed training parameters are provided in \autoref{tab:CycleGAN_setup}.

\begin{table}[!ht]
    \centering
    \small
    \caption{CycleGAN Training Parameters.}
    \label{tab:CycleGAN_setup}
    \begin{tabular}{l|l}
        \toprule
        \textbf{Parameter} & \textbf{Value} \\
        \midrule
        batch\_size & 8 \\
        \midrule
        crop\_size & 256 \\
        \midrule
        load\_size & 286 \\
        \midrule
        gan\_mode & lsgan \\
        \midrule
        netG & resnet\_9blocks \\
        \midrule
        netD & basic \\
        \midrule
        $\beta_1$ & 0.5 \\
        \midrule
        lr & 0.0002 \\
        \midrule
        lr\_policy & linear \\
        \midrule
        Total Epochs & 150 \\ 
        \midrule
        $\lambda_A$, $\lambda_B$ & 10.0 \\
        \midrule
        $\lambda_{identity}$ & 0.5 \\
        \midrule
        norm & instance \\
        \midrule
        preprocess & resize\_and\_crop \\
        \midrule
        num\_threads & 4 \\
        \bottomrule
    \end{tabular}
\end{table}

\subsection{Results and Analysis}

\subsubsection{CycleGAN Style Transfer Results}
\begin{figure*}[t]
    \centering
    \begin{subfigure}[t]{0.23\textwidth}
        \centering
        \includegraphics[width=\textwidth]{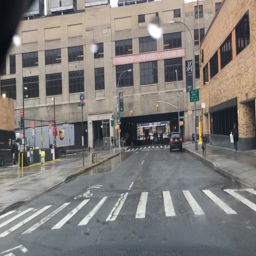}
        \caption{Original Image 1}
        \label{fig:raw1}
    \end{subfigure}
    \hfill
    \begin{subfigure}[t]{0.23\textwidth}
        \centering
        \includegraphics[width=\textwidth]{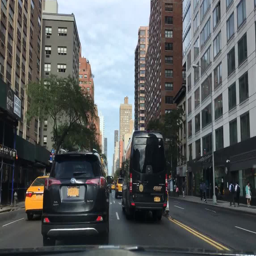}
        \caption{Original Image 2}
        \label{fig:raw2}
    \end{subfigure}
    \hfill
    \begin{subfigure}[t]{0.23\textwidth}
        \centering
        \includegraphics[width=\textwidth]{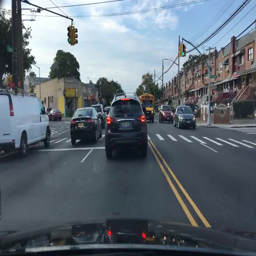}
        \caption{Original Image 3}
        \label{fig:raw3}
    \end{subfigure}
    \hfill
    \begin{subfigure}[t]{0.23\textwidth}
        \centering
        \includegraphics[width=\textwidth]{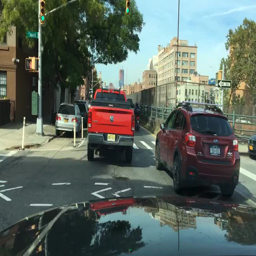}
        \caption{Original Image 4}
        \label{fig:raw4}
    \end{subfigure}
 
    \vspace{0.3cm}
 
    \begin{subfigure}[t]{0.23\textwidth}
        \centering
        \includegraphics[width=\textwidth]{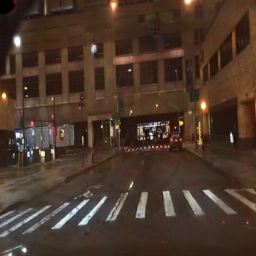}
        \caption{Generated Image 1}
        \label{fig:gen1}
    \end{subfigure}
    \hfill
    \begin{subfigure}[t]{0.23\textwidth}
        \centering
        \includegraphics[width=\textwidth]{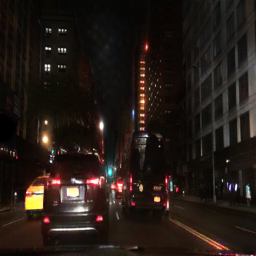}
        \caption{Original Image 2}
        \label{fig:gen2}
    \end{subfigure}
    \hfill
    \begin{subfigure}[t]{0.23\textwidth}
        \centering
        \includegraphics[width=\textwidth]{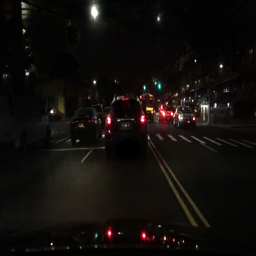}
        \caption{Generated Image 3}
        \label{fig:gen3}
    \end{subfigure}
    \hfill
    \begin{subfigure}[t]{0.23\textwidth}
        \centering
        \includegraphics[width=\textwidth]{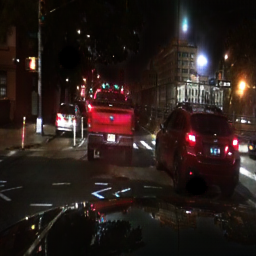}
        \caption{Generated Image 4}
        \label{fig:gen4}
    \end{subfigure}
 
    \caption{Image comparison: The first row (\subref{fig:raw1})-(\subref{fig:raw4}) shows the original daytime images, and the second row (\subref{fig:gen1})-(\subref{fig:gen4}) shows the corresponding generated nighttime images.}
    \label{fig:image_comparison_grid}
\end{figure*}
As shown in \autoref{fig:image_comparison_grid}, the CycleGAN model effectively translates daytime images into nighttime scenes. The model successfully converts daytime features, such as clear lighting and bright skies (Figures \subref{fig:raw1}-\subref{fig:raw4}), into characteristic nighttime visuals, including low-light conditions, road reflections, and illuminated vehicle taillights (Figures \subref{fig:gen1}-\subref{fig:gen4}). Importantly, key semantic elements like vehicle shapes, road structures, and traffic light positions are well-preserved without significant distortion or loss. The generated nighttime images maintain good clarity, allowing for the clear identification of vehicle contours and road markings, which provides a high-quality data foundation for subsequent object detection tasks.

\subsubsection{Data Shift Detection Results}
To evaluate the effectiveness of our CycleGAN-based data augmentation strategy in mitigating data shift, we performed MMD tests between the original daytime training data and the nighttime validation set, as well as between the augmented training data and the nighttime validation set. The average p-values are presented in \autoref{tab:drift_detection_p_value}.

\begin{table}[!ht]
    \centering
    \caption{Comparison of Data Shift Detection Results.}
    \label{tab:drift_detection_p_value}
    \begin{tabular}{c|c}
        \toprule
        \textbf{Dataset Comparison} & \textbf{Average p-value} \\
        \midrule
        $D_{\text{train}}^{\text{base}}$ vs. $D_{\text{val\_night}}$ & 0.03 \\ 
        \midrule
        $D_{\text{train}}^{\text{aug}}$ vs. $D_{\text{val\_night}}$ & 0.15 \\ 
        \bottomrule
    \end{tabular}
\end{table}

The results in \autoref{tab:drift_detection_p_value} show that after CycleGAN-based augmentation, the p-value from the MMD test increased significantly from 0.03 to 0.15. In statistical terms, a higher p-value indicates weaker evidence for a significant difference between the two distributions. This result demonstrates that the distribution of the CycleGAN-augmented training data is significantly more similar to that of the nighttime validation set, effectively reducing the data shift and creating a more favorable condition for the object detection task.

\subsubsection{Impact of Data Shift on YOLOv5s Performance}
\begin{figure}[!ht]
\centering
\begin{subfigure}[b]{0.98\linewidth}
\centering
\includegraphics[width=\linewidth]{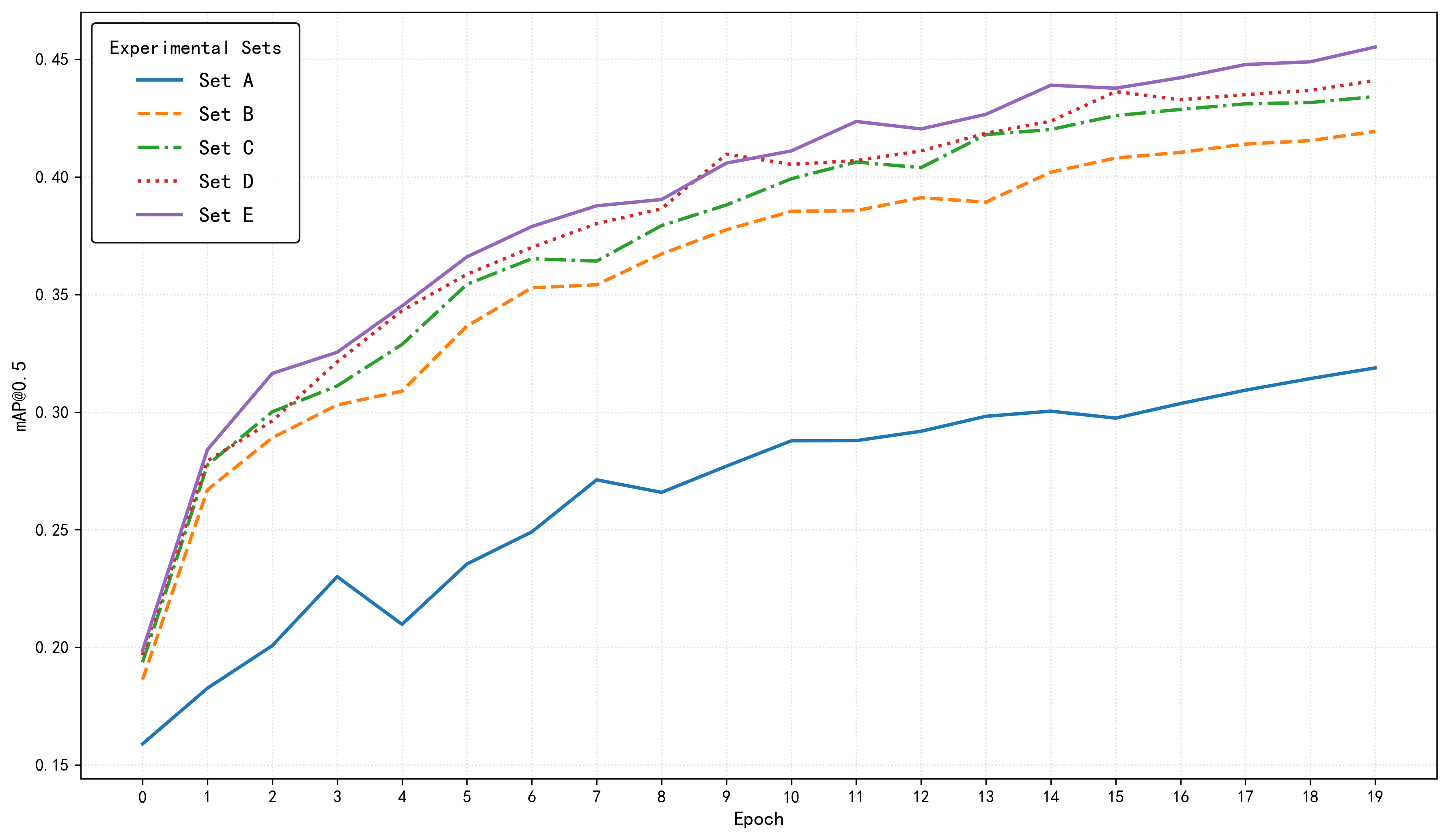}
\caption{mAP@0.5 Trend without Data Augmentation}
\label{fig:no_augmentation_map}
\end{subfigure}
\hfill
\begin{subfigure}[b]{0.98\linewidth}
\centering
\includegraphics[width=\linewidth]{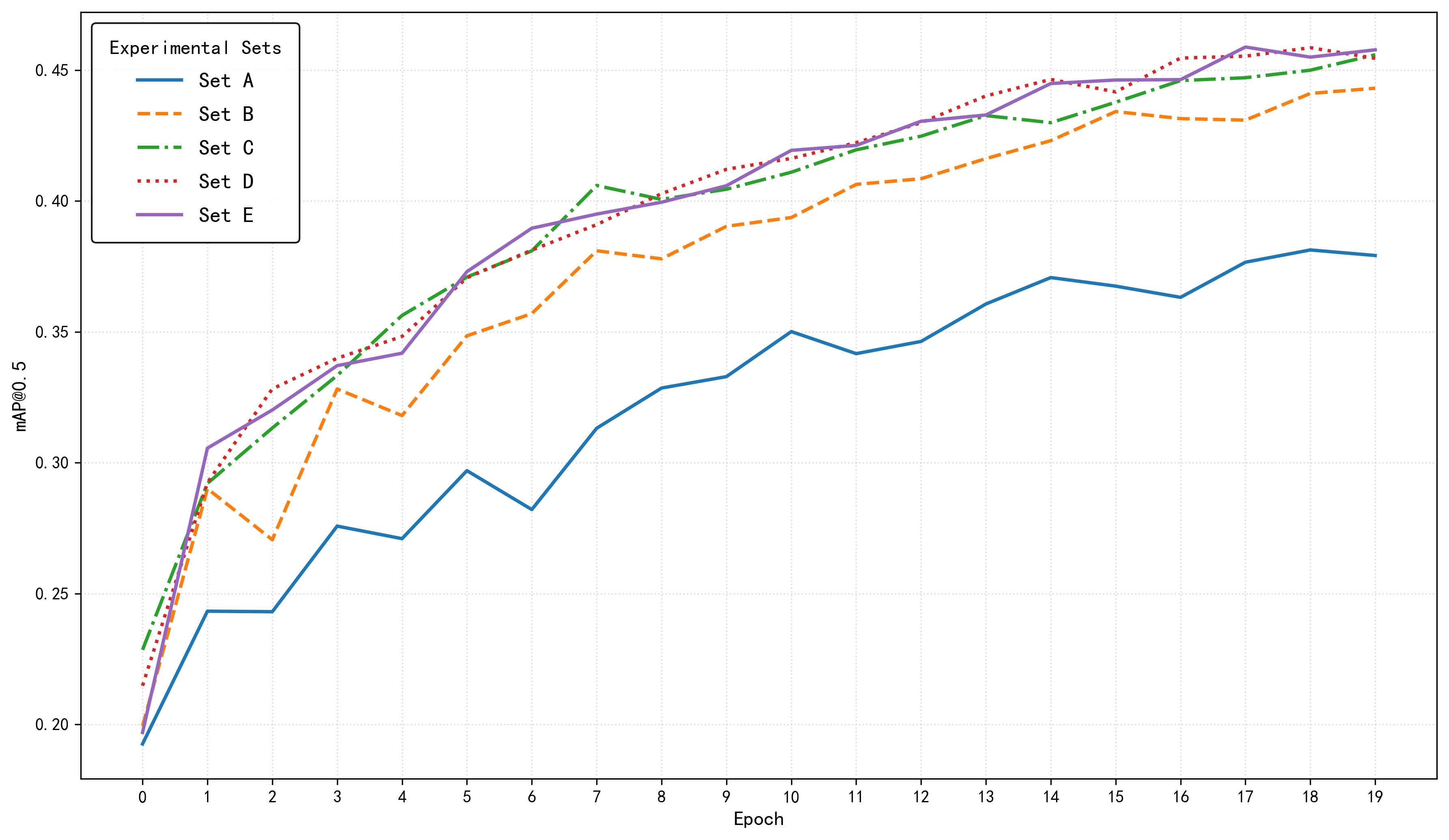}
\caption{mAP@0.5 Trend with Data Augmentation}
 \label{fig:augmentation_map}
\end{subfigure}
\caption{Comparison of YOLOv5s mAP@0.5 learning curves on the night test set for different training configurations.}
\label{fig:map_comparison}
\end{figure}

To investigate the practical impact of data shift and our data augmentation strategy on model performance, we trained YOLOv5s models on the different training sets (Set A to Set E) and evaluated their mean Average Precision (mAP) on the night-only validation set. The results are summarized in \autoref{tab:map_vs_shift} and visualized in \autoref{fig:map_comparison}. \autoref{tab:map_vs_shift} presents the final mAP@0.5 with and without data augmentation, while \autoref{fig:map_comparison} illustrates the mAP@0.5 learning curves throughout the training process.

\begin{table}[!ht]
\centering
\caption{YOLOv5s mAP@0.5 on the Night Validation Set for Different Training Sets and Augmentation Strategies.}
\label{tab:map_vs_shift}
\begin{tabular}{c|c|c|c}
\toprule
 \textbf{Training Set} & \textbf{\makecell{Day/Night \\ Ratio (D/N)}} & \textbf{\makecell{mAP@0.5 \\ (\%)}} & \textbf{\makecell{mAP@0.5 (\%) \\ (w/ Aug.)}} \\
 \midrule
Set A & 40000 / 0 & 31.8 & 37.9 \\
\midrule
Set B & 35000 / 5000 & 41.9 & 44.3 \\
\midrule
Set C & 30000 / 10000 & 43.4 & 45.5 \\
\midrule
Set D & 25000 / 15000 & 44.1 & 45.4 \\
\midrule
Set E & 20000 / 20000 & 45.5 & 45.7 \\
\bottomrule
\end{tabular}
\end{table}

As seen in \autoref{fig:no_augmentation_map}, the model trained exclusively on daytime data (Set A, blue line) exhibits extremely poor generalization to nighttime scenes. Its mAP@0.5 starts at the lowest point, increases slowly, and converges to a final value of only 31.8\%, which is far below the performance of models trained with any amount of nighttime data. This confirms that the covariate shift between day and night domains has a significant negative impact on model performance. As real nighttime data is introduced and its proportion increases (from Set B to Set E), the mAP@0.5 on the night validation set steadily improves. The performance of Set D (44.1\%) and Set E (45.5\%) is notably better than Set A, but the gains begin to diminish as more night data is added. This suggests a phenomenon of diminishing marginal returns, where the model may have already learned the key features of nighttime scenes, and further additions of similar data yield smaller improvements.

Comparing \autoref{fig:no_augmentation_map} (without augmentation) and \autoref{fig:augmentation_map} (with augmentation), it is clear that our data augmentation strategy leads to a consistent upward shift in the mAP@0.5 curves for all training sets and a more stable convergence. As shown in \autoref{tab:map_vs_shift}, the performance of the model trained on Set A improved dramatically from 31.8\% to 37.9\% mAP@0.5 after augmentation. This highlights the effectiveness of our CycleGAN-based approach, especially in scenarios with extreme data shift.

\begin{figure}[!ht]
\centering
\includegraphics[width=0.98\linewidth]{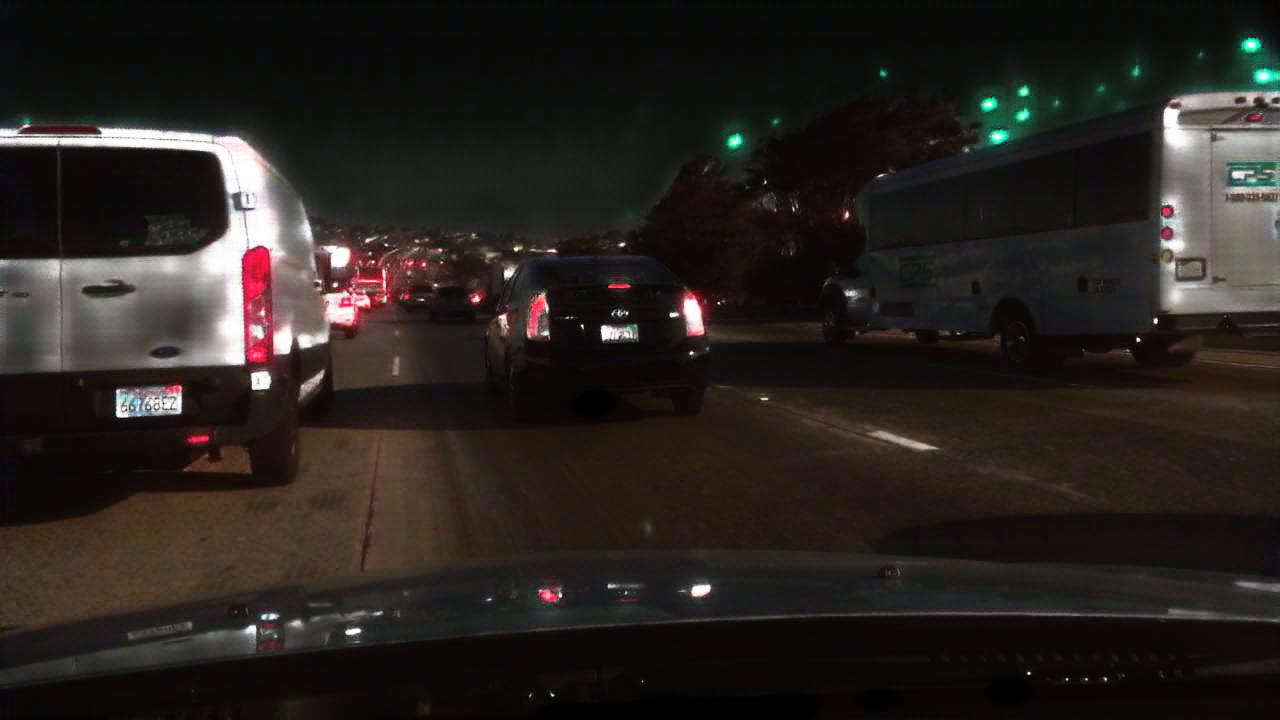} 
\caption{Example of a generated image with noticeable artifacts (green noise in the upper right corner).}
\label{fig:gan_artifacts}
\end{figure}

However, the CycleGAN-based augmentation method also has limitations. As observed in \autoref{fig:augmentation_map} and \autoref{tab:map_vs_shift}, the performance gain from augmentation diminishes as the proportion of real nighttime data in the training set increases. For Set E, the improvement is marginal (45.5\% to 45.7\%). Furthermore, images generated by CycleGAN can sometimes contain unnatural textures and artifacts. For example, \autoref{fig:gan_artifacts} shows a generated image with significant green noise in the sky region. Such artifacts introduce a discrepancy between the augmented data and real nighttime scenes, which can ultimately limit the final accuracy of the object detector.

\section{Conclusion and Future Work}
\label{sec:conclusion}

This paper successfully investigated the significant impact of data shift on the performance of object detection models for autonomous driving and proposed an effective mitigation strategy. By simulating a typical covariate shift using the BDD100K dataset, our research clearly demonstrated that a model trained in a source domain (daytime) suffers a sharp performance degradation when applied to a target domain (nighttime). This underscores the critical importance of enhancing model generalization to unseen or low-frequency environments in real-world autonomous driving applications. To address this challenge, we designed and validated a data augmentation method based on CycleGAN-powered image style transfer, which translates daytime images into a nighttime style to enrich the training set. Experimental results strongly support the efficacy of this strategy, showing that it can significantly boost model performance and effectively alleviate performance drops caused by domain shift, particularly when real data from the target domain is scarce. Additionally, this study leveraged a data shift detection method (MMD) to quantitatively assess the discrepancy between generated and real data distributions, offering a way to anticipate the need for model retraining and potentially save significant computational resources.

Despite the progress made, this study has several limitations. Our research primarily focused on covariate shift, with insufficient exploration of other complex types of data shift, such as concept shift or label shift. Moreover, while CycleGAN-based augmentation is highly effective when target domain data is lacking, its marginal utility diminishes as more real target data becomes available. In such cases, the synthetic data's lack of perfect realism may even introduce minor negative effects. 
This suggests a need for more intelligent, adaptive data augmentation or training strategies, such as online weighted learning methods that can dynamically adjust to the real-time degree of data shift.
In the future, it could explore online or incremental data shift detection methods combined with online or lifelong learning paradigms~\cite{wang2024comprehensive} to enable models to perceive and adapt to new data distributions in real-time \cite{lan2022time}. 
We believe that through continued research and innovation in addressing the data shift problem, we can further advance the maturity and widespread application of autonomous driving technology.

\bibliographystyle{IEEEtran}
\bibliography{bibliography}

\end{document}